\documentclass{article}
\usepackage{spconf,amsmath,graphicx}

\usepackage{enumerate}
\usepackage{amsmath, amssymb}
\usepackage{graphicx}
\graphicspath{ {figures/}}
\usepackage{algorithm}
\usepackage[noend]{algpseudocode}
\usepackage{multirow}
\usepackage{tabularx}
\usepackage{color}

\let\OLDthebibliography\thebibliography
\renewcommand\thebibliography[1]{
  \OLDthebibliography{#1}
  \setlength{\parskip}{0.0pt}
  \setlength{\itemsep}{1.8pt plus 0.3ex}
}

\DeclareMathOperator{\sign}{sign}
\DeclareMathOperator{\top_K_select}{top\_K\_select}
\DeclareMathOperator{\forward}{forward}
\DeclareMathOperator{\backward}{backward}
\DeclareMathOperator{\update}{update}
\DeclareMathOperator{\supp}{supp}

\title{DEEP LEARNING SPARSE TERNARY PROJECTIONS \\ FOR COMPRESSED SENSING OF IMAGES}
\name{Duc Minh Nguyen, Evaggelia Tsiligianni, Nikos Deligiannis}
\address{Vrije Universiteit Brussel, Pleinlaan 2, B-1050 Brussels, Belgium\\
imec, Kapeldreef 75, B-3001 Leuven, Belgium\\
Email: \{mdnguyen, etsiligi, ndeligia\}@etrovub.be}
\begin{document}
\ninept
\maketitle
\begin{abstract}
Compressed sensing (CS) is a sampling theory that
allows  reconstruction of sparse (or compressible) signals 
from an incomplete number of measurements,
using of a sensing mechanism implemented 
by an appropriate \emph{projection matrix}.
The CS theory is based on random Gaussian projection matrices, 
which satisfy recovery guarantees with high probability;
however, sparse ternary $\{0, -1, +1\}$ projections 
are more suitable for hardware implementation.
In this paper, we present a deep learning approach to obtain 
very sparse ternary projections for compressed sensing. 
Our deep learning architecture jointly learns 
a pair of a projection matrix and a reconstruction operator in an end-to-end fashion. 
The experimental results on real images demonstrate the effectiveness of the proposed approach 
compared to state-of-the-art methods, with significant advantage in terms of complexity.
\end{abstract}
\section{Introduction}
\label{sec:intro}

Compressed sensing or compressive sampling (CS) is a theory \cite{Donoho2006, CRT2006_1} 
that merges compression and acquisition, exploiting sparsity to recover signals
that have been sampled at a drastically lower rate than what the Shannon/Nyquist theorem imposes.
The results of CS have an important impact on numerous signal processing applications
including the efficient processing and analysis of high-dimensional
data such as audio \cite{Griffin2008}, image \cite{Gan2007}
and video \cite{mota2016adaptive}.


Assume a finite-length, real-valued signal $x \in \mathbb{R}^{n}$,
CS yields a compressed representation of the treated signal 
using a sensing mechanism that is realized by a \emph{sensing} or \emph{projection matrix}.
The linear measurement process is described by 
\begin{equation}
y = \Phi x,
\label{Psys}
\end{equation}
where $\Phi \in \mathbb {R}^{m \times n}$, $m \ll n$, is the projection matrix and $y \in \mathbb {R}^{m}$ is the vector containing the obtained measurements.
In CS, we assume that either $x$ is a sparse signal
or that $x$ has a sparse representation with respect to a suitable basis  $\Psi\in \mathbb {R}^{n\times n}$,
that is, $x=\Psi u$, $u \in \mathbb {R}^{n}$,  $\|u\|_0=s \ll n$, where $\|\cdot\|_0$ is the $\ell_0$ quasi-norm 
counting the non-vanishing coefficients of the treated signal.
Therefore, we obtain the underdetermined linear system
\begin{equation}
y = \Phi\Psi u.
\label{CSsys}
\end{equation}
A sparse vector satisfying \eqref{CSsys} can be obtained 
as the solution of the $\ell_1$-minimization problem
\begin{equation}
\begin{aligned}
&& \min_{u \in {\mathbb R}^{N}}\left\|u\right\|_{1} \quad  \text{subject to} \quad y=\Phi\Psi u,
\end{aligned}
\label{probleml1CS}
\end{equation}
employing well-known algorithms like Basis Pursuit \cite{ChenDonohoSaunders99}.

Conventional CS theory is based on random Gaussian or random Bernoulli matrices,
which can be used to recover a $n$-dimensional $s$-sparse signal,
provided that the number of measurements $m$ 
is $\mathcal{O}(s\log (n/s))$ \cite{Baraniuk08}.
An important issue when considering random matrices is that
such matrices are typically difficult to build in hardware.
The difficulty in storing these matrices  and certain physical constraints on the measurement process
makes it challenging to realize CS in practice.
Moreover, when multiplying arbitrary matrices with signal vectors of high dimension,
the lack of any fast matrix multiplication algorithm results in high computational cost.
 
Deep learning is an emerging field that learns multiple levels of representation of data
and has been used successfully in image processing tasks.
Existing work has been presented for image super-resolution \cite{Dong2016}, image denoising \cite{Vincent2010} 
and compressed sensing  \cite{Mousavi2015, elad_bcs_arxiv}.
Deep learning has also been applied in distributed CS \cite{Palangi2016},
quantized CS \cite{Sun2016} and video CS \cite{Iliadis2016}.

In this paper, we adopt a deep learning approach
to learn an optimized projection matrix and a non-linear reconstruction mapping
from measurements to the original signal.
In order to design projections suitable for hardware implementation,
we focus on sparse matrices composed of $\{0,-1,+1\}$,
imposing sparsity and binary constraints on the proposed network architecture.
The network is trained on image patches and the learned projection matrix
is used to acquire images in a block-based manner.
Our experimental results show that 
high quality reconstruction can be achieved with
projections containing only $5$\% nonzero $\{-1, 1\}$ entries.

The rest of the paper is organized as follows.
In Section \ref{related} we review related work in CS and deep learning.
Section \ref{proposed} includes the proposed approach
and Section \ref{experiments} our experimental results.
Conclusions are drawn in Section \ref{conclusions}.

\section{Related Work and Our Contributions}
\label{related}

\subsection{Compressed Sensing}

%
While compressed sensing was introduced utilizing random projection matrices,
new results concern matrices that are more efficient
than random projections, leading to fewer necessary measurements
or improvement of the reconstruction performance  \cite{Elad2007, Evaggelia2014}. 
Alternative studies have proposed designs that leverage prior knowledge on the signal~\cite{song2016measurement}
akin to the theory of compressed sensing with prior information~\cite{mota2017compressed}. 
Important research directions focus on
the hardware implementation of CS.
In order to achieve efficient storage, fast encoding and decoding,
structured matrices have been proposed 
\cite{Applebaum2009, Haupt2010}.
Unfortunately, these constructions come at the cost of poor recovery conditions.
Binary and ternary matrices \cite{DeVore2007, Li2014, Amini2011}
yield fast computations;
however, most of the proposed  constructions  are deterministic
and impose restrictions on the matrix dimensions.
Moreover, as it has been proven in  \cite{Baron2010},
LDPC like matrices with entries  $\{0, -1, +1\}$
may perform well as long as they are not too sparse.

Our approach addresses both hardware limitations
and recovery requirements.
The proposed projections can be extremely sparse
(only $5\%$ of their elements are nonzero) allowing highly efficient storage,
and have  $\{-1, +1\}$  nonzero entries
yielding fast computations during acquisition \footnote{The source code of the proposed method is available at https://github.com/nmduc/deep-ternary}. While such a sparse projection matrix would lead to unacceptable recovery performance
when combined with conventional reconstruction algorithms,
the joint optimization of the sensing mechanism and the reconstruction process
yields state-of-the-art results in image recovery.

\subsection{Deep Learning}

The first work concerning recovery from compressed measurements via deep learning is presented in \cite{Mousavi2015}.
The authors use stacked denoising autoencoders 
to jointly train a non-linear sensing operator and a non-linear reconstruction mapping.
While the reconstruction quality is comparable to state-of-the-art algorithms,
the gain in reconstruction time is considerable.
Deep learning for CS has also been employed in \cite{elad_bcs_arxiv}.
Our approach follows similar principles as \cite{Mousavi2015, elad_bcs_arxiv},
but focuses on simplifying the projection matrix. 
While the projection matrices in \cite{Mousavi2015, elad_bcs_arxiv} are dense matrices with real-valued entries, 
we enforce the projection matrix to be a sparse matrix with elements in $\{0, -1, +1\}$
so that it can be efficiently implemented. 

The proposed algorithm is based on recent work on simplifying deep neural networks,
applied on image classification tasks, in which deep networks have achieved great progress.
In \cite{binary_connect},  neural networks are trained with binary weights  $\{-1, +1\}$.  
The study in \cite{binarized_nn_arxiv} extends \cite{binary_connect} to full binary neural networks (BNNs), 
with binary weights and binary hidden unit activations. 
A different technique \cite{xnor_net} adds scaling factors to compensate for the loss introduced by weight binarization. 
%
Another direction in simplifying deep neural networks is to compress pre-trained networks. 
The method proposed in  \cite{efficient_nn_nips2015}  not only learns weights but also connections, 
producing sparsely-connected networks. Extending \cite{efficient_nn_nips2015}, 
in \cite{deep_compression}, connection pruning, weight quantization and Huffman coding
are employed to compress deep neural networks.

Our algorithm incorporates ideas from both these two directions. 
Nevertheless, compared to existing methods, our novelty is two fold: 
(i) We propose a sparsifying technique on the weights implementing the sensing layer;
combined with binarization techniques, our method yields a highly sparse ternary projection matrix.
The learned projections can be stored efficiently and allow fast computations during acquisition,
therefore, they are suitable for hardware implementation.
(ii) We only simplify the first layer in our network, which corresponds to the linear projection matrix 
and allow the reconstruction module to be non-linear in order to achieve high performance.

\section{Learning  Sparse Ternary Projections}
\label{proposed}
Next, we present our proposed method for efficient compressed sensing of images. 
The section starts with a description of our network architecture, followed by our proposed training algorithm.
%
\subsection{Network Architecture}
%
\begin{figure*}[t]
\centering
\includegraphics[width=0.75\textwidth]{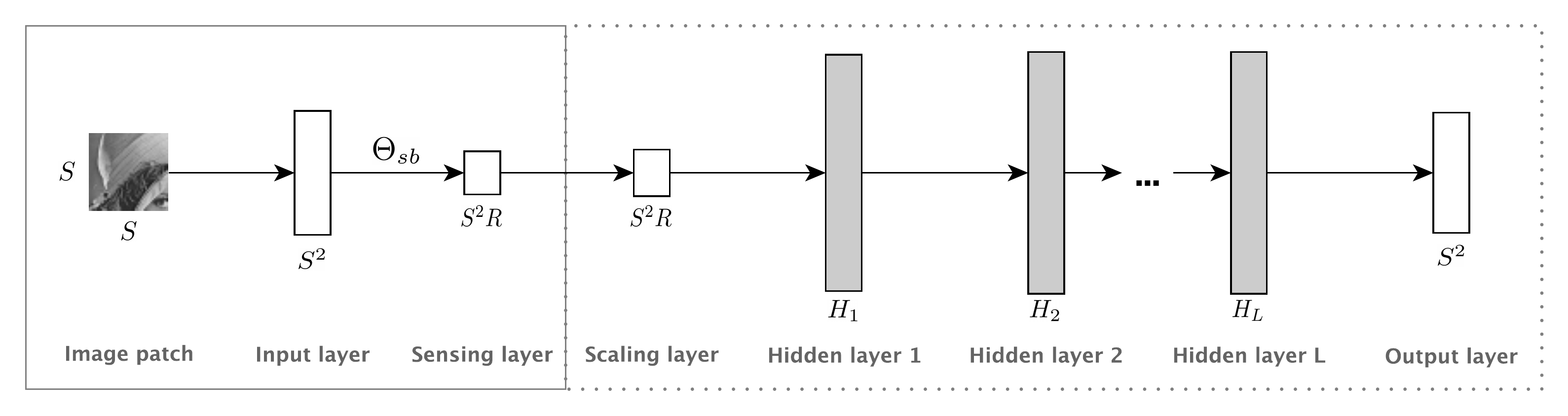}
\caption{Our network architecture: the left block corresponds to the sensing module while the right block (in dash) corresponds to reconstruction module. 
$\Theta_{sb}$ stands for the sparse binary sensing weights. $H_{\ell}$, $\ell = 1, \dots, L$ denotes the number of units of layer $\ell$. 
White blocks denote linear layers while shaded blocks denote non-linear layers.}
\label{fig:architecture}
\end{figure*}
Our network architecture is illustrated in Fig. \ref{fig:architecture}.
It consists of a sensing and a reconstruction module.
The network takes vectorized image patches of size $n=S^{2}$ as input.
The sensing layer projects the $n$-dimensional input signal $x$ to the $m$-dimensional domain;
thus, the number of units in this layer is $m=S^{2}R$, where $0 < R < 1$ is the sensing rate.
The sensing layer has weights $\Theta_{sb} \in \{-1,0,+1\}^{n\times m}$,  
corresponding to the projection matrix $\Phi$, $\Theta_{sb}=\Phi^{T}$.
In order to learn a simple linear projection matrix, 
we do not put bias and non-linearity activation function in this layer. 

The first layer in the reconstruction module is a scaling layer, which
linearly scales the outputs of the sensing layer by learned factors $\alpha$. 
This layer consists of $m$ hidden units connected ``1-1" to the $m$ hidden units 
of the sensing layer and it is employed to compensate for the loss induced by 
binarizing the sensing weights, as explained in Sec. \ref{sec:algorithm}. 
Nevertheless, in deployment, only the projection matrix is implemented in sensing devices 
and the learned scaling factors are kept in the reconstruction module.
The scaling layer is followed by $L$ hidden layers. 
These hidden layers employ the Rectified Linear Unit (ReLU) activation function \cite{relu}. 
The output layer is a linear fully connected layer with size equal to the input dimension. 
All layers in the reconstruction module, except for the scaling layer, are fully connected,  and 
they are followed by a batch-normalization layer \cite{batch_norm}. 
%
\subsection{Training Algorithm}
\label{sec:algorithm}
%
In general, our network training follows the standard mini-batch gradient descent method. 
Denote $x_i, \hat{x}_i$  the input and reconstructed patches, respectively, with $x_i, \hat{x}_i \in \mathbb{R}^n$,  $n=S^{2}$. 
We employ the mean squared error between the input and the reconstruction as our loss function:
\begin{align}
	L = \dfrac{1}{N}\sum_{i=1}^{N} \| x_{i} - \hat{x}_{i} \|_{2}^{2},
\end{align}
where $N$ is the number of sample patches.
In constrast to the conventional mini-batch gradient descent method, 
we introduce a sparsifying and a binarization step 
on the training of the sensing layer. 

More particularly, we first sparsify the continuous-valued sensing weights  $\Theta \in \mathbb{R}^{n\times m}$, 
to get the sparse weights $\Theta_{s} \in \mathbb{R}^{n\times m}$. 
For this step, we propose to retain in $\Theta_{s}$ only the entries
that correspond to the top-K largest absolute values in $\Theta$
and set all the rest entries to zero.
We refer to this procedure as $top\_K\_select$ function. 
The selection of the top weights can be applied column-wise, 
row-wise or over the whole matrix. Nevertheless, since the update of the scaling layer, presented below, involves 
column-wise operator on the sensing weights, we opt to perform $top\_K\_select$ in a column-wise manner.
Each neuron in the sensing layer is connected to $K = S^{2}\gamma$ elements in the input signal, 
where $\gamma$ is the sparsity ratio.
Implementation-wise, we construct a sparse binary mask $M \in \{0,1\}^{n\times m}$  
with entries equal to $1$ corresponding to the largest weights in $\Theta$.
The sparse sensing weights are updated according to $\Theta_{s} = M \odot \Theta$,
where $\odot$ represents the Hadamard product. 

The binarization step involves a mapping of the sparse continuous valued weights  $\Theta_{s} \in \mathbb{R}^{n\times m}$  
to sparse binary weights  $\Theta_{sb} \in \{-1,0,+1\}^{n\times m}$.
Nevertheless, both the sparse and the binarization step introduce some loss, which we want to recover during reconstruction.
For this reason, the sensing layer is followed by a scaling layer similar to the one proposed in \cite{xnor_net}.
This layer with weights $\alpha \in \mathbb{R}^m$ serves as an inverse mapping of the sparse binarized weights $\Theta_{sb}$
to the continuous weights $\Theta$.
Thus, the output $\alpha \Theta_{sb}^T x$ of the scaling layer 
is an approximation of $\Theta^T x$, i.~e., the measurements corresponding to the continuous projections $\Theta^T$.
Let $\theta_{(j)}$ and $\theta_{sb(j)}$ be the $j^{th}$ columns of $\Theta$ and $\Theta_{sb}$, respectively.
$\theta_{(j)}$ corresponds to the dense continuous weights of the $j^{th}$ hidden unit in the sensing layer.
We approximate $\theta_{(j)}$ with $\alpha_{j} \theta_{sb(j)}$, 
where $\alpha_{j} \in \mathbb{R}^{+}$ is a scale factor, corresponding to the $j^{th}$ entry of the scaling weights $\alpha$.
The values of $\theta_{sb(j)}$ and $\alpha_{j}$ can be determined by minimizing the following mean square error with respect to $\theta_{sb(j)}$, $\alpha_{j}$:
\begin{equation}
	E = \| \theta_{(j)} - \alpha_j \theta_{sb(j)} \|_{2}^{2}.
	\label{eq:approx}
\end{equation}
By expanding (\ref{eq:approx}), we have:
\begin{equation}
	E = \theta_{(j)}^{T}\theta_{(j)} - 2 \alpha_j \theta_{(j)}^{T} \theta_{sb(j)} + \alpha_{j}^{2} \theta_{sb(j)}^{T} \theta_{sb(j)},
	\label{eq:approx_expand}
\end{equation}
where $\theta_{(j)}^{T}\theta_{(j)}$ is a constant. 
As $\alpha_j$ is a positive scalar, following \cite{xnor_net} 
but also taking into account our sparsity constraint,  
we obtain the optimal sparse binary vector $\theta_{sb(j)}$ as a solution to the problem:
%
\begin{align}
	\theta_{sb(j)}^{*} = &\underset{\theta_{sb(j)}}{\mathrm{argmax}}(\theta_{(j)}^{T}\theta_{sb(j)}), \nonumber\\
	& \text{s.t.} \quad \theta_{sb(j)} \in \{-1, 0, +1\}^{n} \nonumber \\
	& \qquad \supp(\theta_{sb(j)}) = \supp (\theta_{s(j)}),
\label{eq:prob_theta_sb}
\end{align}
%
where  $\theta_{s(j)}$ is the $j^{th}$ column of $\Theta_s$, and $\supp (\cdot)$ denotes the positions of the nonzero entries of the treated vector. 
The solution of (\ref{eq:prob_theta_sb})  is a vector containing the signs of $\theta_{s(j)}$.
After obtaining the optimal $\theta_{sb(j)}$, we can solve for the optimal $\alpha_{j}$ 
by making the derivative of $E$ equal to zero. 
Considering that $\theta_{sb(j)}^{T} \theta_{sb(j)} = K$, the optimal $\alpha_{j}$ is given by: 
\begin{equation}
	\alpha_{j} =  \dfrac{1}{K} \theta_{(j)}^{T} \theta_{sb(j)} =  \dfrac{1}{K}\sum_{i=1}^{n}\theta_{(ji)} \theta_{sb(ji)} = \dfrac{1}{K}\|\theta_{s(j)}\|_{1}.
	\label{eq:update_alpha_1}
\end{equation}
After the sparsifying and binarization steps, the resulting $\Theta_{sb}$ is sparse and has $K$ nonzero entries in $\{-1, +1\}$ in each of its column.

Following existing training algorithms for networks with binary weights \cite{binary_connect,binarized_nn_arxiv,xnor_net}, 
we use the sparse binary weights $\Theta_{sb}$ during forward and backward propagation. 
The high precision weights $\Theta$, on the other hand, are used during parameter update, to accommodate the small changes of the weights after each update step. 
In our training, $\Theta$ is updated using the gradient of the loss function with respect to $\Theta_{sb}$. 
It should be noted that even though $\Theta_{sb}$ contains only discrete weights in $\{0, -1, +1\}$, 
the gradient of the loss with respect to it still lies in the continuous domain.
Denoting $\Theta^{t}, \Theta^{t}_{s}, \Theta^{t}_{sb}$  the continuous, sparse, and sparse binary sensing weights, respectively,
$\alpha^{t}$ the scaling layer's weights, $W^{t}$ the reconstruction weights 
and $\theta_{s(j)}^{t}$ the $j^{th}$ column of $\Theta_s^{t}$ at step $t$,
our training is summarized by Algorithm \ref{alg:training}.

\begin{algorithm}[t]
\caption{The proposed training algorithm of sparse ternary projection matrix and reconstruction weights at step $t$.}
\label{alg:training}
\textbf{Input:} The patches $X$; the weights $\Theta^{t-1}$ and $W^{t-1}$, the learning rate $\mu$ \\
\textbf{Output:} The loss $L$; the updated weights $\Theta^{t}$, $W^{t}$; the sparse binary weights $\Theta_{sb}^{t}$; the scaling weights $\alpha^{t}$
\begin{algorithmic}[1]
\Procedure{Sparsify and binarize sensing weights}{}
\State $M \gets \top_K_select(\Theta^{t-1})$
\State $\Theta_{s}^{t} \gets M \odot \Theta^{t-1}$
\State $\Theta_{sb}^{t} \gets \sign(\Theta_{s}^{t})$
\State $\alpha^{t}_{j} \gets \dfrac{1}{K}||\theta^{t-1}_{s(j)}||_{1} , \quad \forall j \in [1,m]$
\EndProcedure
\Procedure{Forward propagation}{}
\State $L \gets \forward(X, \Theta_{sb}^{t}, W^{t-1})$
\EndProcedure
\Procedure{Backward propagation}{}
\State $\dfrac{\delta L}{\delta \Theta_{sb}^{t}}, \dfrac{\delta L}{\delta W^{t-1}} \gets \backward(L, \Theta_{sb}^{t}, W^{t-1})$
\EndProcedure
\Procedure{Parameter update}{}
\State $W^{t} \gets \update(W^{t-1}, \dfrac{\delta L}{\delta W^{t-1}}, \mu)$
\item[]
\State $\Theta^{t} \gets \update(\Theta^{t-1}, \dfrac{\delta L}{\delta \Theta_{sb}^{t}}, \mu)$
\EndProcedure
\end{algorithmic}
\end{algorithm}

\section{Experimental results}
\label{experiments}
In order to evaluate the proposed algorithm, we carried out experiments in image recovery.
We employed the ILSVRC2012 validation set \cite{ILSVRC15}, including $50$ K images for training, 
and tested our model on two testing sets of images of resolution $256\times 256$. 
The first testing set consists of 10 images, taken from the ILSVCR2014 \cite{ILSVRC15} dataset, 
and is provided by the authors of \cite{Mousavi2015}. 
The second testing set is composed of 50 images randomly selected from the LabelMe dataset \cite{labelme}.
All the images were converted to grayscale in our experiments. 
To reduce the computational overhead, we ran our experiments in small image patches of size $32\times 32$ pixels. 
In total, we randomly sampled $5$ millions patches to form our training set. 
During training, the input patches were preprocessed by subtracting the mean and dividing by the standard deviation. 
We trained our network using Algorithm \ref{alg:training}, with the Adam parameter update \cite{adam}, 
a batch size of $5000$, $50$ epochs and a learning rate of $0.01$ decaying by a factor of $0.6$ every $5$ epochs. 
The training samples were randomly shuffled after each epoch. 
To avoid over-fitting, we employ $\ell_2$ regularization on the reconstruction modules, with a weight equal to $0.001$.
During the testing phase, we sampled overlapping patches from each test image, with a stride of $2$ pixels 
and determined the final image reconstruction as the average of the patches' reconstructions. 
The methods are evaluated using PSNR values, expressed in dB.
Concerning the network architecture, we empirically set the number of non-linear hidden layers $L$ to $2$,
each with $2048$ hidden units since this configuration produces a good trade-off between training time and reconstruction quality.
\\
\noindent \\
\textbf{Sensing rate} \\
\noindent
First, we experiment with different sensing rates. We use $\gamma = 0.1$ and vary $R$ in $[0.1,0.3]$. 
The mean PSNR values on the first testing set are shown in Table \ref{table:exp2_sensing_rate}.
As shown in the table, the overall reconstruction quality gets better with larger 
sensing rates, since more information from the signal is retained in the measurements.
\begin{table}[t]
\centering
\caption{Reconstruction performance when varying $R$ ($\gamma = 0.1$).}
\label{table:exp2_sensing_rate}
\begin{tabular}{| c | c | c | c | c | c |}
\hline
$R$ & $0.1$ & $0.15$ & $0.2$ & $0.25$ & $0.3$ \\
\hline
PSNR & 25.24 & 26.11 & 26.65 & 27.52 & 27.96 \\
\hline
\end{tabular}
\end{table}
\\
\noindent \\
\textbf{Sparsity ratio} \\
\noindent
Next, we experiment with different sparsity ratios, using $R=0.25$ and varying $\gamma$. 
The mean PSNR values on the first testing set are presented in Table \ref{table:exp3_gamma}.
The dimension of input is $32\times 32 $ and, for $R=0.25$, we obtain $\Theta_{sb} \in \{-1,0,+1\}^{1024\times 256}$.
With $\gamma \in \{0.001, 0.005, 0.010, 0.050, 0.100, 0.300, 1.000\}$, the number of nonzero entries in each column of $\Theta_{sb}$ (i.~e. $K$) is $1$, $5$, $10$, $51$, $102$, $307$, $1024$, respectively.
As can be seen, acceptable reconstruction performance can be achieved using extremely sparse projection matrices with only $0.1$\% nonzero entries ($\gamma=0.001$). 
Varying $\gamma$ from $0.001$ to $0.005$ considerably improves the performance, while the difference between $\gamma=0.005$ and $\gamma=0.010$ is negligible. 
The network reaches its peak performance with $\gamma=0.050$ and performs slightly worse with $\gamma \in \{0.100, 0.300, 1.000\}$.
It should be pointed out that with $\gamma \in \{0.001, 0.005\}$, there are $256$ and $1280$ nonzero entries in the projection matrix, respectively. 
The former is not enough to fully cover the $1024-$dimensional input signal. 
We argue that this is the reason for the noticeable performance jump when increasing $\gamma$ to $0.005$.
During training, the network experiences over-fitting with $\gamma \in \{0.1, 0.3, 1.0\}$. This explains why $\gamma=0.050$ 
gives better performance than $\gamma \in \{0.1, 0.3, 1.0\}$. 
As a result, the proposed sparse binary constraint can be considered as an extra regularizer to the network.
\begin{table}[t]
\centering
\caption{Reconstruction performance when varying $\gamma$ ($R = 0.25$).}
\label{table:exp3_gamma}
\begin{tabular}{| c | p{0.07\linewidth} | p{0.07\linewidth} | p{0.07\linewidth} | p{0.07\linewidth} | p{0.07\linewidth} | p{0.07\linewidth} | p{0.07\linewidth} |}
\hline
$\gamma$ & $0.001$ & $0.005$  & $0.010$ & $0.050$ & $0.100$ & $0.300$ & $1.000$ \\
\hline
PSNR & 25.83 & 26.96 & 26.98 & 27.61 & 27.52 & 27.40  & 27.37 \\
\hline
\end{tabular}
\end{table}
\\
\noindent \\
\textbf{Comparison with state of the art} \\
\noindent
As the proposed algorithm implements CS via deep learning, the next experiment involves a comparison with the method of \cite{Mousavi2015},
which employs a stacked denoising autoencoder to jointly learn the sensing layer and the reconstructor. 
We select the best algorithm from \cite{Mousavi2015}, referred to as O-NL-SDA for our comparison. 
This algorithm uses a non-linear sensing mechanism, with overlapping image patches of size $32 \times 32$.
The results of O-NL-SDA on the first testing set is taken from \cite{Mousavi2015}.
To obtain the results on the second testing set, we train an O-NL-SDA model on our training set using the proposed configurations in \cite{Mousavi2015}.
Results obtained with a conventional reconstruction algorithm, namely, Basis Pursuit (BP) \cite{ChenDonohoSaunders99}, using random ternary projections are also presented.
Sparse binary and ternary constructions like the ones proposed in \cite{Li2014, Amini2011} could not be employed in our experiments
due to the constraints they impose on matrix dimensions. 
To have a fair comparison with \cite{Mousavi2015}, we use the same sensing rate, $R=0.25$. 
We choose $\gamma=0.050$ for the proposed method since it yields the best performance, while producing a highly sparse projection matrix.
The comparison between the selected methods on the first testing set is shown in Table \ref{table:exp4_compare_stoa_oi}. On the second testing set, the mean PSNR values for O-NL-SDA \cite{Mousavi2015}, BP \cite{ChenDonohoSaunders99} and the proposed algorithm are $31.38$, $24.04$ and $33.15$ dB, respectively.
%
%
\begin{table}[t]
\centering
\caption{Reconstruction performance (PSNR) of different algorithms on the first testing set.}
\label{table:exp4_compare_stoa_oi}
\begin{tabular}{| c | c | c | c |}
\hline
& O-NL-SDA \cite{Mousavi2015} & BP \cite{ChenDonohoSaunders99} & Proposed \\
\hline
Damselfly		& 30.85	& 26.49	& 31.56 \\ \hline
Birds		    & 26.62	& 22.75	& 27.17	\\ \hline
Rabbit			& 27.24	& 22.63	& 27.89	\\ \hline
Turtle			& 34.65	& 26.31	& 35.59	\\ \hline
Dog				& 21.55	& 16.97	& 22.33	\\ \hline
Eagle Ray		& 26.57	& 22.65	& 27.20	\\ \hline
Boat		    & 33.11	& 25.48	& 33.94	\\ \hline
Monkey			& 30.32	& 23.51	& 30.70	\\ \hline
Panda			& 21.00	& 17.85	& 21.53	\\ \hline
Snake			& 17.71	& 14.67	& 18.23	\\ \hline
\textit{Mean PSNR} & \textit{26.96} & \textit{21.93} & \textit{27.61} \\ \hline
\end{tabular}
\end{table}
%
%
As can be seen, the proposed algorithm yields significantly better results than
the conventional reconstruction with BP \cite{ChenDonohoSaunders99} and ternary but not sparse projections.
Despite having a sparse ternary matrix of only 5\% of nonzero entries, our method outperforms 
O-NL-SDA \cite{Mousavi2015} in terms of the recovery performance. 
Concerning the speed of the reconstructor, 
as it was reported in \cite{Mousavi2015}, a reconstructor implemented using 
a feed-forward neural network can perform orders of magnitude faster than a convex optimization solver. 
Clearly, our method can provide a convenient hardware implementation of the sensing mechanism
and a fast reconstructor with better reconstruction quality than the state of the art.

\section{Conclusions}
\label{conclusions}
In this paper, we propose a novel algorithm to train a pair of a highly sparse ternary projection matrix and a reconstruction operator 
for compressed sensing of images. 
The sparse and ternary structure of the learned projection matrix can be exploited in efficient hardware implementations. 
Experimental results on real images  show that the achieved reconstruction performance
for a projection matrix with $5$\% nonzero binary entries and a corresponding reconstructor trained end-to-end with the proposed algorithm
outperforms state-of-the-art methods.
Extremely sparse projection matrices with only $0.1$\% nonzero entries learned using the same algorithm
yield acceptable performance as well.

\subsection*{Acknowledgment}
We would like to thank the authors of \cite{Mousavi2015} for providing us the test data
and the code implementing the related method. The research has been supported by Fonds Wetenschappelijk Onderzoek (project no. G0A2617) 
and Vrije Universiteit Brussel (PhD bursary Duc Minh Nguyen, research programme M3D2).

\bibliographystyle{IEEEbib}
\bibliography{refs}

\begin{thebibliography}{10}

\bibitem{Donoho2006}
D.~L. Donoho,
\newblock ``Compressed sensing,''
\newblock {\em IEEE Trans. Inf. Theory}, vol. 52, no. 4, pp. 1289--1306, 2006.

\bibitem{CRT2006_1}
E.~J. Cand\`{e}s, J.~Romberg, and T.~Tao,
\newblock ``Robust uncertainty principles: Exact signal reconstruction from
  highly incomplete frequency information,''
\newblock {\em IEEE Trans. Inf. Theory}, vol. 52, no. 2, pp. 489--509, 2006.

\bibitem{Griffin2008}
A.~Griffin and P.~Tsakalides,
\newblock ``Compressed sensing of audio signals using multiple sensors,''
\newblock in {\em European Signal Processing Conference (EUSIPCO)}, 2008, pp.
  1--5.

\bibitem{Gan2007}
L.~Gan,
\newblock ``Block compressed sensing of natural images,''
\newblock in {\em International Conference on Digital Signal Processing
  (ICDSP)}, 2007, pp. 403--406.

\bibitem{mota2016adaptive}
J.~F.~C. Mota, N.~Deligiannis, A.~C. Sankaranarayanan, V.~Cevher, and M.~R.~D.
  Rodrigues,
\newblock ``Adaptive-rate reconstruction of time-varying signals with
  application in compressive foreground extraction,''
\newblock {\em IEEE Trans. Signal Process.}, vol. 64, no. 14, pp. 3651--3666,
  2016.

\bibitem{ChenDonohoSaunders99}
S.~S. Chen, D.~L. Donoho, and M.~A. Saunders,
\newblock ``Atomic decomposition by basis pursuit,''
\newblock {\em SIAM J. Sci. Comput.}, vol. 20, no. 1, pp. 33--61, 1999.

\bibitem{Baraniuk08}
R.~Baraniuk, M.~Davenport, R.~DeVore, and M.~Wakin,
\newblock ``A simple proof of the restricted isometry property for random
  matrices,''
\newblock {\em Constr. Approx}, vol. 28, no. 3, pp. 253--263, 2008.

\bibitem{Dong2016}
C.~Dong, C.~C. Loy, K.~He, and X.~Tang,
\newblock ``Image super-resolution using deep convolutional networks,''
\newblock {\em IEEE Trans. Pattern Anal. Mach. Intell.}, vol. 38, no. 2, pp.
  295--307, 2016.

\bibitem{Vincent2010}
P.~Vincent, H.~Larochelle, I.~Lajoie, Y.~Bengio, and P.~Manzagol,
\newblock ``{Stacked Denoising Autoencoders}: Learning useful representations
  in a deep network with a local denoising criterion,''
\newblock {\em J. Mach. Learn. Res.}, vol. 11, pp. 3371--3408, Dec. 2010.

\bibitem{Mousavi2015}
A.~Mousavi, A.~B. Patel, and R.~G. Baraniuk,
\newblock ``A deep learning approach to structured signal recovery,''
\newblock in {\em Annual Allerton Conference on Communication, Control, and
  Computing (Allerton)}, 2015, pp. 1336--1343.

\bibitem{elad_bcs_arxiv}
A.~Adler, D.~Boublil, M.~Elad, and M.~Zibulevsky,
\newblock ``A deep learning approach to block-based compressed sensing of
  images,''
\newblock {\em ArXiv e-prints}, June 2016.

\bibitem{Palangi2016}
H.~Palangi, R.~Ward, and L.~Deng,
\newblock ``Exploiting correlations among channels in distributed compressive
  sensing with convolutional deep stacking networks,''
\newblock in {\em IEEE International Conference on Acoustics, Speech and Signal
  Processing (ICASSP)}, 2016, pp. 2692--2696.

\bibitem{Sun2016}
B.~Sun, H.~Feng, K.~Chen, and X.~Zhu,
\newblock ``A deep learning framework of quantized compressed sensing for
  wireless neural recording,''
\newblock {\em IEEE Access}, vol. 4, pp. 5169--5178, 2016.

\bibitem{Iliadis2016}
M.~Iliadis, L.~Spinoulas, and A.~K. Katsaggelos,
\newblock ``{DeepBinaryMask}: Learning a binary mask for video compressive
  sensing,'' http://arxiv.org/abs/1607.03343,
\newblock 2016.

\bibitem{Elad2007}
M.~Elad,
\newblock ``Optimized projections for compressed sensing,''
\newblock {\em IEEE Trans. Signal Process.}, vol. 55, no. 12, pp. 5695--5702,
  2007.

\bibitem{Evaggelia2014}
E.~Tsiligianni, L.~P. Kondi, and A.~K. Katsaggelos,
\newblock ``{Construction of Incoherent Unit Norm Tight Frames with Application
  to Compressed Sensing},''
\newblock {\em IEEE Trans. Inf. Theory}, vol. 60, no. 4, pp. 2319--2330, 2014.

\bibitem{song2016measurement}
P.~Song, J.~F.~C. Mota, N.~Deligiannis, and M.~R.~D. Rodrigues,
\newblock ``Measurement matrix design for compressive sensing with side
  information at the encoder,''
\newblock in {\em Statistical Signal Processing Workshop (SSP)}. IEEE, 2016,
  pp. 1--5.

\bibitem{mota2017compressed}
J.~F.~C. Mota, N.~Deligiannis, and M.~R.~D. Rodrigues,
\newblock ``Compressed sensing with prior information: Strategies, geometry,
  and bounds,''
\newblock {\em IEEE Trans. Inf. Theory}, vol. 63, no. 7, pp. 4472--4496, 2017.

\bibitem{Applebaum2009}
L.~Applebaum, S.~D. Howard, S.~Searle, and R.~Calderbank,
\newblock ``Chirp sensing codes: Deterministic compressed sensing measurements
  for fast recovery,''
\newblock {\em App.\ Comp.\ Harm.\ Anal.}, vol. 26, no. 2, pp. 283--290, 2009.

\bibitem{Haupt2010}
J.~D. Haupt, W.~U. Bajwa, G.~Raz, and R.~Nowak,
\newblock ``Toeplitz compressed sensing matrices with applications to sparse
  channel estimation,''
\newblock {\em IEEE Trans. Inf. Theory}, vol. 56, no. 11, pp. 5862--5875, 2010.

\bibitem{DeVore2007}
R.~A. DeVore,
\newblock ``Deterministic constructions of compressed sensing matrices,''
\newblock {\em J. Complexity}, vol. 23, no. 4-6, pp. 918--925, 2007.

\bibitem{Li2014}
S.~Li and G.~Ge,
\newblock ``Deterministic construction of sparse sensing matrices via finite
  geometry,''
\newblock {\em IEEE Trans. Signal Process.}, vol. 62, no. 11, pp. 2850--2859,
  2014.

\bibitem{Amini2011}
A.~Amini and F.~Marvasti,
\newblock ``Deterministic construction of {Binary}, {B}ipolar and {T}ernary
  compressed sensing matrices,''
\newblock {\em IEEE Trans. Inf. Theory}, vol. 57, no. 4, pp. 2360--2370, 2011.

\bibitem{Baron2010}
D.~Baron, S.~Sarvotham, and R.~G. Baraniuk,
\newblock ``Bayesian compressive sensing via belief propagation,''
\newblock {\em IEEE Trans. Signal Process.}, vol. 58, no. 1, pp. 269--280,
  2010.

\bibitem{binary_connect}
M.~Courbariaux, Y.~Bengio, and J-P. David,
\newblock ``Binaryconnect: Training deep neural networks with binary weights
  during propagations,''
\newblock in {\em International Conference on Neural Information Processing
  Systems (NIPS)}, 2015, pp. 3123--3131.

\bibitem{binarized_nn_arxiv}
M.~Courbariaux, I.~Hubara, D.~Soudry, R.~El-Yaniv, and Y.~Bengio,
\newblock ``Binarized neural networks: Training deep neural networks with
  weights and activations constrained to +1 or -1,''
\newblock {\em ArXiv e-prints}, Feb. 2016.

\bibitem{xnor_net}
M.~Rastegari, V.~Ordonez, J.~Redmon, and A.~Farhadi,
\newblock ``X{NOR}-net: Imagenet classification using binary convolutional
  neural networks,''
\newblock in {\em European Conference on Computer Vision (ECCV)}, 2016, pp.
  525--542.

\bibitem{efficient_nn_nips2015}
S.~Han, J.~Pool, J.~Tran, and W.~J. Dally,
\newblock ``Learning both weights and connections for efficient neural
  networks,''
\newblock in {\em International Conference on Neural Information Processing
  Systems (NIPS)}, 2015, pp. 1135--1143.

\bibitem{deep_compression}
S.~Han, H.~Mao, and W.~J. Dally,
\newblock ``Deep compression - compressing deep neural networks with pruning,
  trained quantization and huffman coding,''
\newblock in {\em International Conference on Learning Representations (ICLR)},
  2016.

\bibitem{relu}
X.~Glorot, A.~Bordes, and Y.~Bengio,
\newblock ``Deep sparse rectifier neural networks,''
\newblock in {\em International Conference on Artificial Intelligence and
  Statistics (AISTATS)}, 2011, pp. 315--323.

\bibitem{batch_norm}
S.~Ioffe and C.~Szegedy,
\newblock ``Batch normalization: Accelerating deep network training by reducing
  internal covariate shift,''
\newblock in {\em International Conference on Machine Learning (ICML)}, 2015,
  pp. 448--456.

\bibitem{ILSVRC15}
O.~Russakovsky, J.~Deng, H.~Su, J.~Krause, S.~Satheesh, S.~Ma, Z.~Huang,
  A.~Karpathy, A.~Khosla, M.~Bernstein, A.~C. Berg, and L.~Fei-Fei,
\newblock ``Imagenet large scale visual recognition challenge,''
\newblock {\em Int. J. Comput. Vis.}, vol. 115, no. 3, pp. 211--252, 2015.

\bibitem{labelme}
B.~C. Russell, A.~Torralba, K.~P. Murphy, and W.~T. Freeman,
\newblock ``Labelme: A database and web-based tool for image annotation,''
\newblock {\em Int. J. Comput. Vis.}, vol. 77, no. 1-3, pp. 157–173, 2008.

\bibitem{adam}
D.~P. Kingma and J.~L. Ba,
\newblock ``Adam: A method for stochastic optimization,''
\newblock in {\em International Conference on Learning Representations (ICLR)},
  2015.

\end{thebibliography}

\end{document}